\documentclass[journal]{IEEEtran}
\usepackage{graphicx}
\usepackage[utf8]{inputenc}
\usepackage{graphics} % for pdf, bitmapped graphics files
\usepackage{epsfig} % for postscript graphics files
\usepackage{mathptmx} % assumes new font selection scheme installed
\usepackage{amsmath} % assumes amsmath package installed
\usepackage{amssymb}  % assumes amsmath package installed
\usepackage{cite}
\usepackage{physics}
\usepackage{multicol}
\usepackage{mathtools, cuted}
\usepackage[cmintegrals]{newtxmath}
\usepackage{epstopdf}
\usepackage{graphics} % for pdf, bitmapped graphics files
\usepackage{epsfig} 

\usepackage{mathptmx} % assumes new font selection scheme installed
\usepackage{amsmath} % assumes amsmath package installed
\usepackage{amssymb}  % assumes amsmath package installed
\usepackage{cite}
\usepackage{multicol}
\usepackage{mathtools, cuted}
\usepackage[cmintegrals]{newtxmath}
\usepackage{mathtools}

\usepackage{mathptmx}
\usepackage{helvet}
\usepackage{courier}

\usepackage{type1cm}
\usepackage{titlesec}
\usepackage{epsfig} % for postscript graphics files
\usepackage{mathptmx} % assumes new font selection scheme installed
\usepackage{amsmath} % assumes amsmath package installed
\usepackage{amssymb}  % assumes amsmath package installed
\usepackage{cite}
\usepackage{multicol}
\usepackage{mathtools, cuted}
\usepackage[cmintegrals]{newtxmath}
\usepackage{makeidx}         % allows index generation
\usepackage{algorithm}% http://ctan.org/pkg/algorithm
\usepackage{algpseudocode}% http://ctan.org/pkg/algorithmicx

\usepackage{multicol}        % used for the two-column index
\usepackage[bottom]{footmisc}% places footnotes at page bottom

\usepackage{nomencl}
\usepackage{url}
\usepackage{siunitx}
\usepackage[utf8]{inputenc}
\usepackage{url}
\usepackage{dblfloatfix}
\usepackage[caption=false,font=footnotesize]{subfig}
\usepackage{hyperref}

\usepackage{multicol}        % used for the two-column index
\usepackage[bottom]{footmisc}% places footnotes at page bottom

\usepackage{titlesec}
\usepackage{nomencl}
\usepackage[usenames, dvipsnames]{color}
\usepackage{multirow}
\usepackage{adjustbox}
\usepackage{amsthm}

\usepackage[table]{xcolor}
\usepackage[normalem]{ulem} % for \sout

\newif\ifrevdraft
\revdrafttrue     % <--- set to \revdraftfalse for a clean (final) PDF

\newcommand{\revcap}[1]{\ifrevdraft\textcolor{blue}{#1}\else#1\fi}

\newenvironment{revblock}{\par\ifrevdraft\color{blue}\fi}{\par}

\theoremstyle{definition}

\theoremstyle{remark}

\begin{document}

\title{VR-Based Control of Multi-Copter Operation\\
% {\footnotesize \textsuperscript{*}Note: Sub-titles are not captured for https://ieeexplore.ieee.org  and
}

\author{Jack T. Hughes,~Garegin Mazmanyan,~ Mohammad Ghufran,  
        and~Hossein Rastgoftar,~\IEEEmembership{Member,~IEEE}% <-this % stops a space
\thanks{Authors are with the Aerospace and Mechanical Engineering Department, University of Arizona, Tucson, AZ, 85719 USA email: \{jath03, gmazmanyan, ghufran1942, hrastgoftar\}@arizona.edu  }% <-this % stops a space
% \thanks{J. Doe and J. Doe are with Anonymous University.}% <-this % stops a space
% \thanks{Manuscript received April 19, 2005; revised August 26, 2015.}
}

% \and
% \and
% \and
% \and
% \and
\markboth{
% IEEE Aerospace and Electronic Systems Magazine
}%
{Shell \MakeLowercase{\textit{et al.}}: Bare Demo of IEEEtran.cls for IEEE Journals}
\maketitle

% \begin{abstract}

% \begin{oldblock}
% We aim to use virtual reality (VR) to improve the spatial awareness of pilots by real-time scanning of the environment around the drone using onboard sensors, live streaming of this environment to a VR headset, and rendering a virtual representation of the drone and its environment for the pilot. This way, the pilot can see the immediate environment of the drone up close from a third-person perspective, as opposed to the first-person perspective that most drone cameras provide. This provides much more information about the drone’s surroundings for the pilot while operating the drone than existing teleoperation solutions. Previous solutions using VR have relied upon pre-made designs of the environment, which makes it difficult to adapt to changing environments. Our solution, in contrast, scans the environment as you fly, making it much more flexible for use in unknown environments. 
% \end{oldblock}

 \begin{revblock}
\textbf{Abstract---}We present a VR-based teleoperation system for multirotor flight that renders a third-person view (TPV) of the vehicle together with a live 3D reconstruction of its surroundings. The system runs on an embedded GPU (Jetson Orin NX) with ROS2/WebXR integration and streams geometry and video to a headset for closed-loop control in previously unmapped spaces. We implement a first-person video (FPV) baseline and perform matched trials with two pilots in unmapped indoor spaces. Quantitative metrics are reported from repeated trials with one pilot ($N{=}8$). TPV achieved task time comparable to FPV while improving proximal obstacle awareness (minimum obstacle distance ${+}0.20\,\mathrm{m}$) and reducing contacts. These results indicate that TPV can preserve control quality while exposing hazards less visible in FPV, supporting safer teleoperation in unknown environments.
\end{revblock}

% \end{abstract}


\section{Introduction}
Although autonomous multi-copter drones have increasingly found real-world applications, human-piloted multi-copters are still widely used in industries such as aerial inspection, photography, and defense.  To date, there have been two primary visual input methods provided to pilots: the first is direct line-of-sight operation \cite{han2022uas, hobbs2016human}, and the second involves the use of onboard front- or downward-facing cameras \cite{sato2025development}. The first method enhances situational awareness by requiring the pilot to maintain unaided visual contact with the aircraft throughout the entire flight, within the Visual Line of Sight (VLOS). As a result, this method limits drone operation to within close proximity of the pilot. In contrast, the second method enables pilots to operate drones beyond the line of sight using onboard cameras. However, it restricts the pilot's perception to the direction the camera is facing, which reduces situational awareness and can lead to crashes when the drone travels in other directions. 

We propose a novel solution that eliminates the major limitations of existing methods. In our system, drone pilots use a virtual reality (VR) headset to view the drone in virtual space from a third-person perspective, as if it were directly in front of them. In addition to the drone itself, a three-dimensional representation of its surroundings is displayed, without requiring any prior knowledge of the drone's environment.

\section{Related Work}
Aerial inspection by multirotor drones has received a lot of attention from many industrial sectors, such as mining \cite{holuvsa2022utilization, kamran2024applications}, agriculture \cite{de2020virtual, huuskonen2018soil}, and construction \cite{cheng2022construction, albeaino2022dronesim}.  Teleoperation is the available tool for controlling multicopters from the ground, where the information collected by onboard sensors can be communicated to a ground station computer. When using such a traditional manual multi-copter control system, the pilot has limited spatial awareness and the ability to communicate control commands to the drone, which means they cannot be adjusted based on in-situ observations the drone gathers. Furthermore, multi-copters' onboard sensors   are usually fixed on the body frame, and therefore, real-time data acquisition quality is significantly affected by the orientation of the aircraft's body frame, and as a result, the data collected by onboard cameras are not necessarily useful and informative. 
% Add literature review

Virtual reality (VR) has greatly enhanced human-robot interaction in drone control by providing immersive interfaces that improve spatial awareness. In the context of drones, traditional  First-Person View (FPV) systems suffer from limited depth perception, while VR offers more intuitive control. In \cite{9974281} and \cite{10.1145/3574131.3574432}, it is demonstrated how VR improves maneuverability and control. Refs. \cite{9981656} and \cite{9341037} highlight how VR can enhance spatial awareness and human-centered interaction, and \cite{CHEN2020105579} shows that using VR aids in operative drones in a complex environment. Recent work on Mixed Reality (MR) interfaces, such as the one designed for omnidirectional aerial vehicles (OMAVs) further enhances the user interaction by allowing operators to control both translational and rotational in six degree freedom \cite{10156426}.

Traditional systems for 3D reconstruction that rely on pre-built maps struggle with real-time updates in dynamic environments. Multi-sensor fusion and real-time SLAM improve adaptability and accuracy in changing environments like urban and off-road \cite{doi:10.1142/S2301385020500168} \cite{9833300} \cite{9635985}. These systems benefit autonomous drones, enabling navigation in unknown spaces, while VR systems rely on real-time mapping for immersive experiences \cite{9526756} \cite{9381521} \cite{app13179877}.

In drone teleportation, spatial orientation and situational awareness are critical. \cite{9981656} highlighted that while FPV is immersive, it often limits the operator's situational awareness by restricting to the drone's camera field of view. In contrast, third-person view (TPV) systems significantly enhance spatial awareness, providing a broader environmental perspective and superior situational awareness \cite{10.1145/3544548.3580681} \cite{temma}. Study on monoscopic and stereoscopic views shows that TPV improves depth perception, aiding in better control\cite{s17081720}.

\begin{revblock}
\noindent\textbf{Third-person limitations.}
Prior work found that third-person viewpoints can degrade navigation performance, reduce spatial orientation, and weaken embodiment in immersive tasks \cite{Medeiros2018_ThirdPersonBadForNavVR}.
We consider a different use case, teleoperation of a physical drone.
Here TPV complements egocentric cues by revealing lateral and rear hazards that FPV often hides.
We therefore evaluate TPV against an FPV baseline and plan a larger study to identify when TPV helps and when it hurts.
\end{revblock}

\begin{figure}[ht]
\centering
\includegraphics[width=0.49 \textwidth]{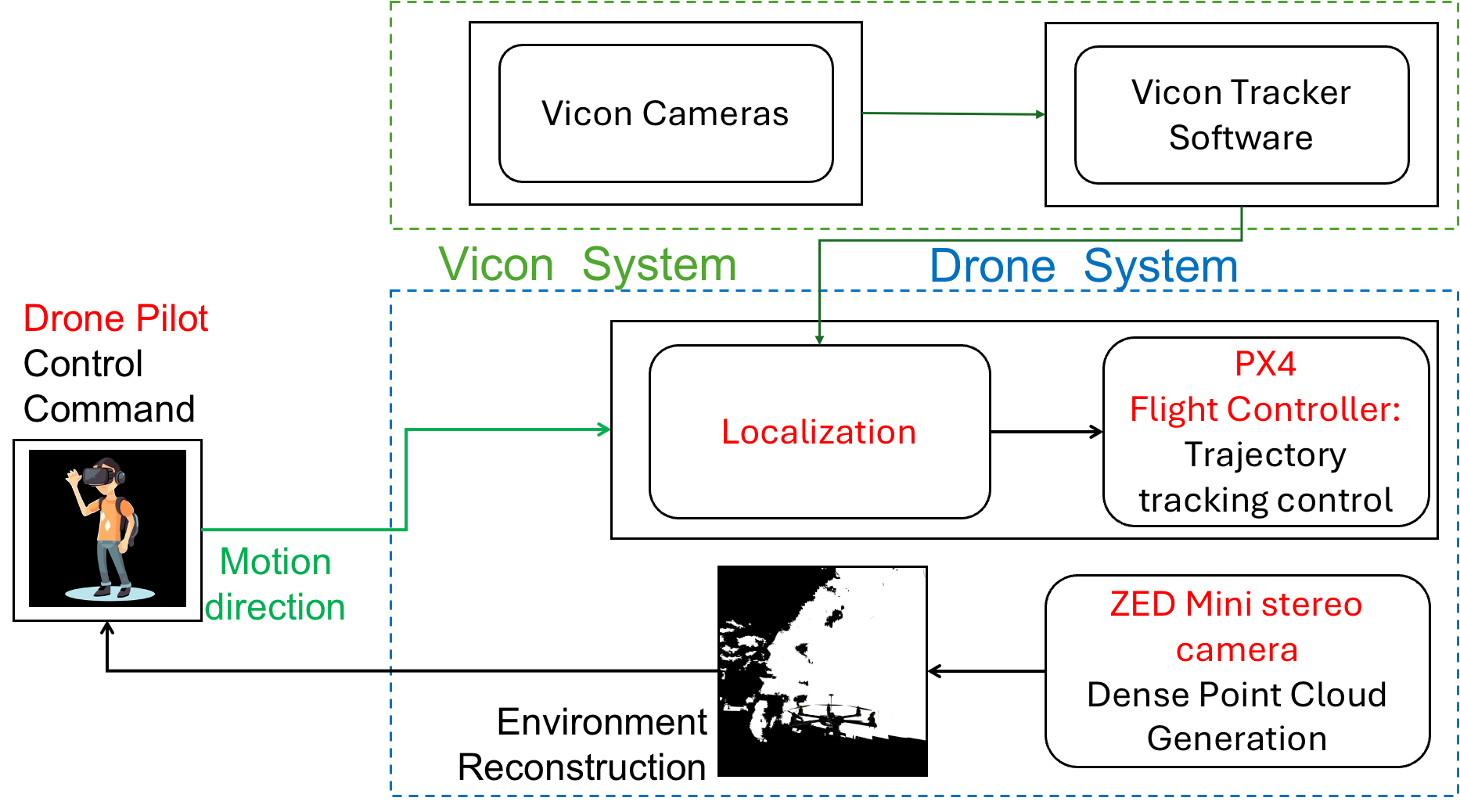}
\vspace{-0.3cm}
\caption{\label{pilo-multicoptercontrol}The architecture proposed for the VR-based perception and control.}
\end{figure}

\subsection{Contributions}
Existing VR systems for drone teleoperation often rely on static environments and FPV, limiting adaptability and situational awareness and struggling with real-time updates and navigating dynamic, unknown spaces. Our research overcomes these gaps by integrating real-time environment scanning with TPV, enhancing adaptability and spatial awareness for efficient drone operation. In particular, we propose to enhance the spatial awareness of drone by providing virtual reality (VR) capabilities to better sense and control drone operation and interaction. 	More specifically, we build a novel architecture shown in Fig. \ref{pilo-multicoptercontrol} for pilot-onboard control of drones using virtual reality to safely operate drones in a constrained  environment. 
 
In our work, a multicopter drone with ZED mini stereo cameras is utilized to map the environment in real time, where the drone pilot interprets the mapping and  authorizes a proper motion direction command based on a predetermined control scheme. This scheme defines the relationships between the drone pilot and her/his corresponding actions, such as takeoff, landing, hovering, pitch, roll, and yaw maneuvers. The drone’s onboard flight controller analyzes the received control signals and adjusts the rotational speed of its motors to perform the desired maneuvers and maintain stable flight. 

For our experiment with the VR-Based control system, we want to verify that our system is a viable option  for controlling multicopters.  To perform this initial experiment, we had a pilot put on the VR headset and, with only our system as visual input, try to fly a multicopter in an unmapped indoor environment towards a wall, stop, return to the starting position, and land safely.  This simple test can demonstrate three important points.  First, it shows that the system provides enough visual feedback about the multicopter’s position and orientation in space to fly it in a controlled manner.  Second, it shows that the system’s latency is low enough that the pilot can react to the multicopter’s movement accordingly. Third, it shows that an unknown environment can be mapped and displayed to the pilot in real time. 
% To ensure the safety of the surroundings, the control system will need to incorporate failsafe mechanisms and emergency protocols, such as gesture validation, signal loss, emergency stop, and obstacle avoidance. We propose, as shown in Figure 3, to implement Thrust 2.

\section{System}
Our system includes a hexacopter, a human operator, and Meta Quest 3 for establishing interaction between the human and hexacopter. The hexacopter model and control are presented in \ref{Quadcopter Model} and \ref{Controller}, respectively, followed by the description of the VR system in Section \ref{meta}.
\subsection{Hexacopter Model}\label{Quadcopter Model}
To localize the hexacopter motion, we define a global coordinate system that is fixed on the ground with orthonormal base vectors $\hat{\mathbf{e}}_1$, $\hat{\mathbf{e}}_2$, and $\hat{\mathbf{e}}_3$. We also define a local (body) coordinate system that is fixed on the hexacopter body, with orthonormal base vectors $\hat{\mathbf{c}}_1$, $\hat{\mathbf{c}}_2$, and $\hat{\mathbf{c}}_3$. We can use the $3-2-1$ Euler angle standard to characterize the orientation of the hexacopter at any time, where we use $\phi$, $\theta$, and $\psi$ to denote the hexacopter's roll, pitch, and yaw angles, respectively. The angular velocity of the hexacopter is expressed with respect to the body coordinate system by 
\begin{equation}
    \mathbf{\Omega}=p\hat{\mathbf{c}}_1+q\hat{\mathbf{c}}_2+r\hat{\mathbf{c}}_3
\end{equation}
where $p$, $q$, and $r$ are the angular velocity components when it is expressed with respect to the body coordinate system and  obtained based on $\phi$, $\theta$, $\psi$, $\dot{\phi}$, $\dot{\theta}$, and $\dot{\psi}$  by 
\begin{equation}
    \begin{bmatrix}
        p\\
        q\\
        r
    \end{bmatrix}
    =\begin{bmatrix}
    1&0&\sin \theta\\
    0&\cos\phi&\cos\theta \sin \phi\\
    0&-\sin\phi&\cos\phi\cos \theta
    \end{bmatrix}
\end{equation}
The equation of motion of a hexacopter with mass $m$ and mass moment of inertia tensor $\mathbf{J}$ is obtained by
\begin{subequations}
    \begin{equation}
        m\ddot{\mathbf{r}}=-mg\mathbf{e}_3+f\hat{\mathbf{c}}_3,
    \end{equation}
    \begin{equation}
        \mathbf{J}\dot{\mathbf{\Omega}}+{\mathbf{\Omega}}\times \mathbf{J}{\mathbf{\Omega}}=\mathbf{T}
    \end{equation}
\end{subequations}
where $\mathbf{r}$ is the position of the hexacopter measured with respect to the inertial coordinate system, $f$ is the magnitude of the thrust force generated by all six rotors, and  $\mathbf{T}$ is control torque. 

\subsection{Controller}\label{Controller}

The hexacopter applies the  cascaded position and attitude controllers.
% which are cascaded with the output of the first given to the second as input.
The position controller aim to generate desired force
% \vspace{-0.2cm}
\begin{equation}
    \mathbf{F}_{des} = - \mathbf{K}_p \left(\dot{\mathbf{r}} - \dot{\mathbf{r}}_T\right) - \mathbf{K}_v \left(\mathbf{r} - \mathbf{r}_T\right) + mg\hat{\mathbf{e}}_3 + m\ddot{\mathbf{r}}_T,
\end{equation}
where $\mathbf{r}_T$ and $\dot{\mathbf{r}}_T$ are the commanded position and velocity, and $\mathbf{K}_p$ and $\mathbf{K}_v$ are positive definite gain matrices. By projecting $\mathbf{F}_{des}$ onto  $\hat{\mathbf{c}}_3$, the desired thrust force is denoted by $f_{des}$ and obtained by
\vspace{-0.2cm}
\begin{equation}
    f_{des} = \mathbf{F}_{des}.\hat{\mathbf{c}}_3,
\end{equation}
By knowing $\mathbf{F}_{des}$, the desired orientations of the body base vectors are denoted by $\hat{\mathbf{c}}_{1,des}$, $\hat{\mathbf{c}}_{2,des}$, and $\hat{\mathbf{c}}_{3,des}$,  and obtained by
\begin{subequations}
\begin{equation}\label{eq_16}
    \hat{\mathbf{c}}_{3,des} = \frac{\mathbf{F}_{des}}{\lVert\mathbf{F}_{des}\rVert}.
\end{equation}
\label{eq:base}
\begin{equation}
\label{eq:i_B}
    \hat{\mathbf{c}}_{1,des} = \frac{\left(\hat{\mathbf{e}}_3 \times \hat{\mathbf{h}}\right) \times \hat{\mathbf{c}}_{3,des}}{\left|\left|\left(\hat{\mathbf{e}}_3 \times \hat{\mathbf{h}}\right) \times \hat{\mathbf{c}}_{3,des}\right|\right|},
\end{equation}
\begin{equation}
    \hat{\mathbf{c}}_{2,des}  = \hat{\mathbf{c}}_{3,des} \times \hat{\mathbf{c}}_{1,des}.
\end{equation}
\end{subequations}
% \vspace{-0.2cm}
% Note that \eqref{eq_16} assume that $\lVert\mathbf{F}_{des}\rVert \neq 0$.
where
\begin{equation}
\label{eq:heading_def}
   \hat{\mathbf{h}} =  \frac{(\hat{\mathbf{c}}_{1}\cdot\hat{\mathbf{e}}_{1})\hat{\mathbf{e}}_{1} + (\hat{\mathbf{c}}_{1}\cdot\hat{\mathbf{e}}_{2})\hat{\mathbf{e}}_2 }{||(\hat{\mathbf{c}}_{1}\cdot\hat{\mathbf{e}}_{1})\hat{\mathbf{e}}_{1} + (\hat{\mathbf{c}}_{1}\cdot\hat{\mathbf{e}}_{2})\hat{\mathbf{e}}_2 ||}.
\end{equation}
is the heading vector. By knowing  $\hat{\mathbf{c}}_{1,des}$, $\hat{\mathbf{c}}_{2,des}$, and $\hat{\mathbf{c}}_{3,des}$, we obtain desired rotation matrix 
\begin{equation}
    \mathbf{S}_{des}=\begin{bmatrix}
        \hat{\mathbf{c}}_{1,des}&\hat{\mathbf{c}}_{2,des}&\hat{\mathbf{c}}_{3,des}
    \end{bmatrix}
\end{equation}
We also define rotation matrix 
\begin{equation}
    \mathbf{S}=\begin{bmatrix}
        \hat{\mathbf{c}}_{1}&\hat{\mathbf{c}}_{2}&\hat{\mathbf{c}}_{3}
    \end{bmatrix}
\end{equation}
and compute attitude error as follows:
% \vspace{-0.2cm}
\begin{equation}
    \mathbf{e}_R = \frac{1}{2}\left(\mathbf{S}_{des}^T\mathbf{S}-\mathbf{S}^T\mathbf{S}_{des}\right)^\vee.
\end{equation}
where $\square^\vee$ is the vee map which transforms a skew-symmetric matrix into a vector by
\vspace{-0.2cm}
\begin{equation}
    \begin{bmatrix}0&a&b\\-a&0&c\\-b&-c&0\end{bmatrix}^\vee = \begin{bmatrix}-c\\b\\-a\end{bmatrix}, \qquad \forall (a,b,c)\in \mathbb{R}^3.
\end{equation}

We denote the desired torque that needs to be generated by the hexacopter rotors  by $\mathbf{T}_{des}$ and compute it by
% \vspace{-0.3cm}
\begin{equation}
    \mathbf{T}_{des} = -\mathbf{K}_R\mathbf{e}_R -\mathbf{K}_\omega \mathbf{e}_{\omega},
\end{equation}
where $\mathbf{K}_R$ and $\mathbf{K}_\omega$ are diagonal gain matrices, and 
% We use $\mathbf{\omega}_T$ to compute angular velocity error as defined by:
% \vspace{-0.3cm}
\begin{equation}
    \mathbf{e}_{\omega} = \mathbf{\Omega} - \mathbf{\Omega}_T.
\end{equation}
Note that $\mathbf{\Omega}_T$ is the target angular velocity.

% Hexacopter Model
% \subsection{Control}
\subsection{Meta Quest 3}\label{meta}
Our system uses a Meta Quest 3 to render and display the virtual environment.  The rendering is done in a WebXR-based website through Three.js and is hosted on the onboard Jetson computer.  This enables the system to be used on any VR headset that supports WebXR and even allows for multiple simultaneous users.  The headset is connected to ROS2 through rosbridge\_server, as shown in Fig. \ref{softwaresetup}.

The virtual environment uses the coordinate system created by the ZED camera, where the starting point of the camera, and thus the multi-copter, is used as the origin and the environment is placed relative to that point.  Before rendering, this coordinate system is shifted 5 meters in front of the pilot so that the pilot sees the multi-copter and environment in front of them instead of being co-located.

% Meta Quest 3 -> how the heacopter model and vicon model super imposition

\begin{figure}[ht]
% \centering
\includegraphics[width=0.49 \textwidth]{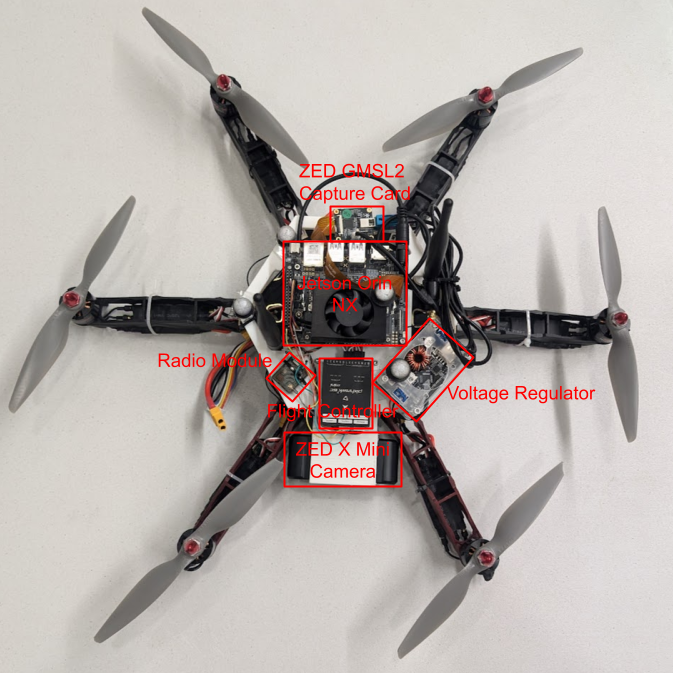}
% \vspace{-0.3cm}
\caption{\label{hardwaresetup}The hexacopter used for the experiment.}
\end{figure}
\section{Experimental Setup}

The experiment is conducted is a robotic facility equipped with 17 Vicon Valkyrie cameras and a Vicon tracking system offering high-resolution tracking system within a  {30}{m} $\times$ {15}{m} $\times$ {5}{m} indoor testing facility.

\subsection{Hardware Setup}

For the experimental validation of our proposed approach, we are utilizing a hexacopter (Fig. \ref{hardwaresetup}) equipped with Nvidia Jetson Orin NX 16GB (Table \ref{table:jetson}) and SeteroLabs ZED X Mini Camera (Table \ref{table:camera}). The Camera is connected to Jetson using a ZED GMSL2 capture card, which processes the camera image using the Jetson GPU. The Jetson and the camera are powered by a 4-cell Lithium Polymer battery, with a voltage regulator stepping down the voltage from 16.8V to 12V to ensure stable power delivery.

\begin{table}[ht]
% \centering
\vspace{-5mm}
\caption{Nvidia Jetson Specifications}
\vspace{-2mm}
\begin{tabular}{r|l}
\textbf{Parameter}          & \textbf{Description} \\ \hline \hline
GPU                         & 1024-core NVIDIA Ampere GPU with 32 Tensor Cores \\ \hline
CPU                         & 8-core Arm® Cortex®-A78AE v8.2 64-bit \\ \hline
Memory                      & 16GB 128-bit LPDDR5 \\ \hline
Storage                     & 128 GB ssd \\ \hline
\end{tabular}
\label{table:jetson}
\end{table}

Our flight controller of choice is the Pixhawk 6C mini, paired with a Flysky radio for remote control and communication. To provide external pose data for flight, we use the Vicon motion capture system, which increases positional accuracy. We designed and 3D-printed custom mounts for both the Jetson and the camera to integrate them securely into the drone's frame.

\begin{table}[ht]
\caption{Camera Specification}
\vspace{-2mm}
\centering
\begin{tabular}{r|l}
\textbf{Parameter}          & \textbf{Description} \\ \hline \hline
Camera Resolution           & 600p at 60 FPS \\ \hline
Field of View (FOV)         & 110 (H) x 80 (V) x 120° (D) max \\ \hline
Depth Range                 & 0.1m to 8m (0.3ft to 26ft) \\ \hline
Dimensions                  & 94 x 32 x 37 mm (3.69 x 1.25 x 1.44 in) \\ \hline
Baseline                    & 5 cm (1.97 in) \\ \hline
Weight                      & 150 g (0.33 lb) \\ \hline
\end{tabular}
\vspace{-3mm}
\label{table:camera}
\end{table}

\begin{figure}[ht]
\centering
\includegraphics[width=0.48 \textwidth]{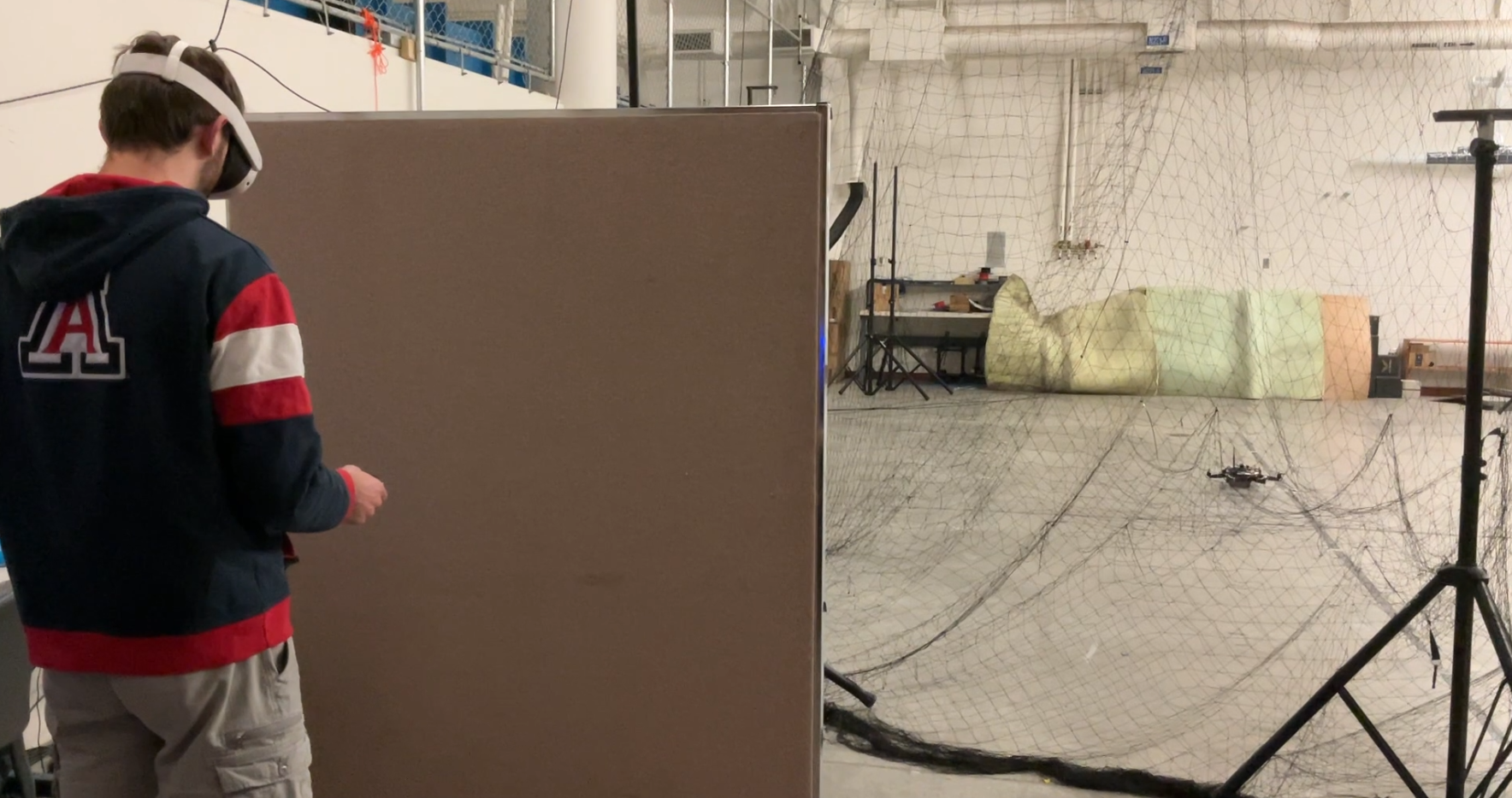}
% \vspace{-0.3cm}
\caption{Operator with Meta Quest 3}
\label{fig:jack_with_headset}
\end{figure}

For user interaction and control, Mixed Reality (MR) provides intuitive holographic controls, enhancing situational awareness, and user studies confirm the effectiveness of MR controls across various experience levels \cite{10156426}. For the mixed reality (MR) component, we are using Meta Quest 3 (Fig \ref{fig:jack_with_headset}) due to its cutting-edge features, such as high-resolution displays, standalone functionality, and mixed realities capabilities.  Control is achieved using the included Meta Quest Touch Plus controllers.
% For the mixed reality (MR) component, we are using Meta Quest 3 due to its cutting-edge features, such as high-resolution displays, standalone functionality, and mixed realities capabilities. 

% \subsubsection{Vicon Motion Capture System}

% \subsubsection{Hexacopter Drone}

\begin{figure}[ht]
\centering
\includegraphics[width=0.48 \textwidth]{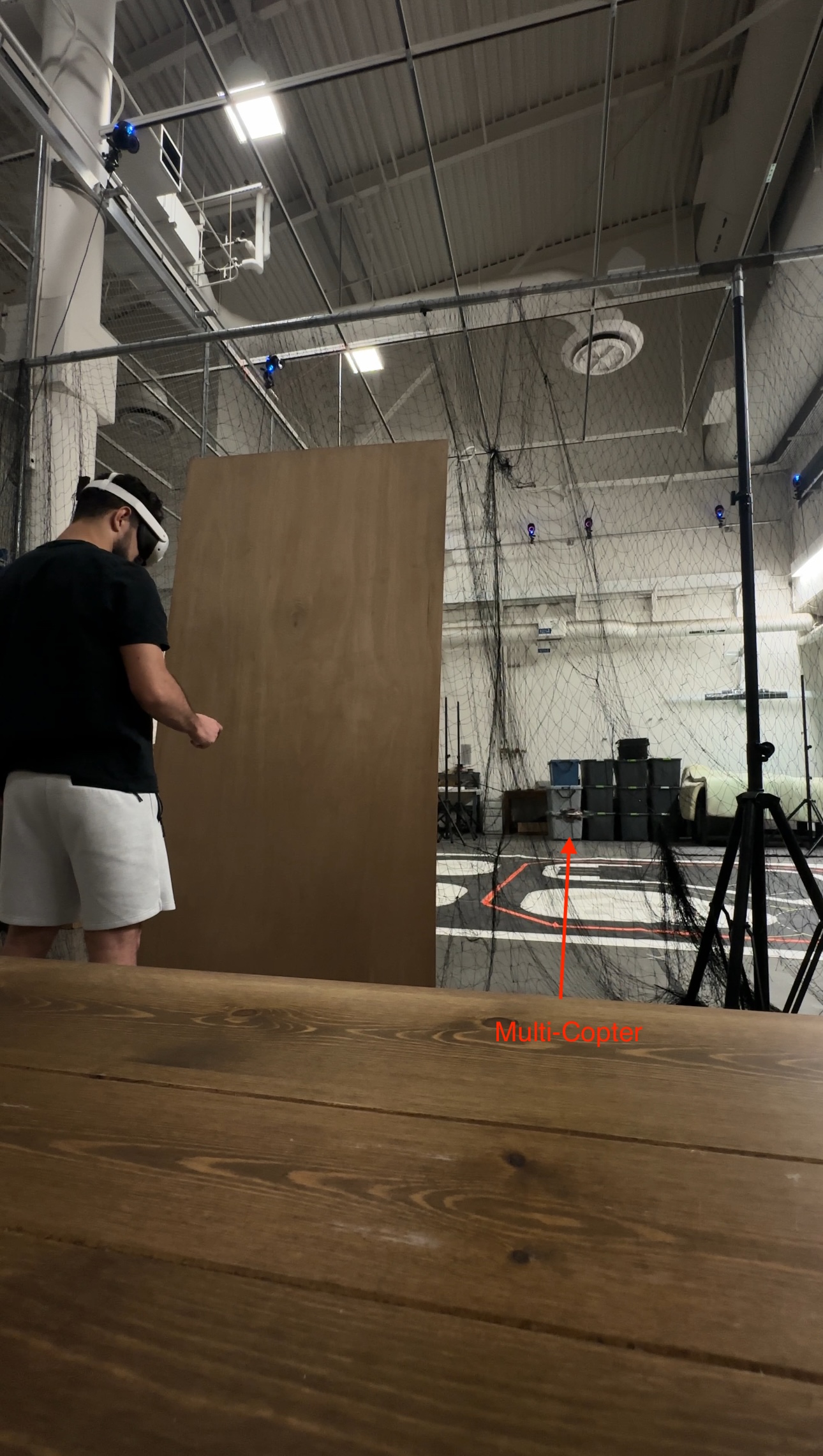}
% \vspace{-0.3cm}
\caption{\revcap{Second pilot conducting VR teleoperation with the Meta Quest~3 inside the safety enclosure (TPV/FPV trials).}}

\end{figure}

\subsection{Software Setup}

\begin{figure}[t]
    \centering
    \includegraphics[width=\linewidth]{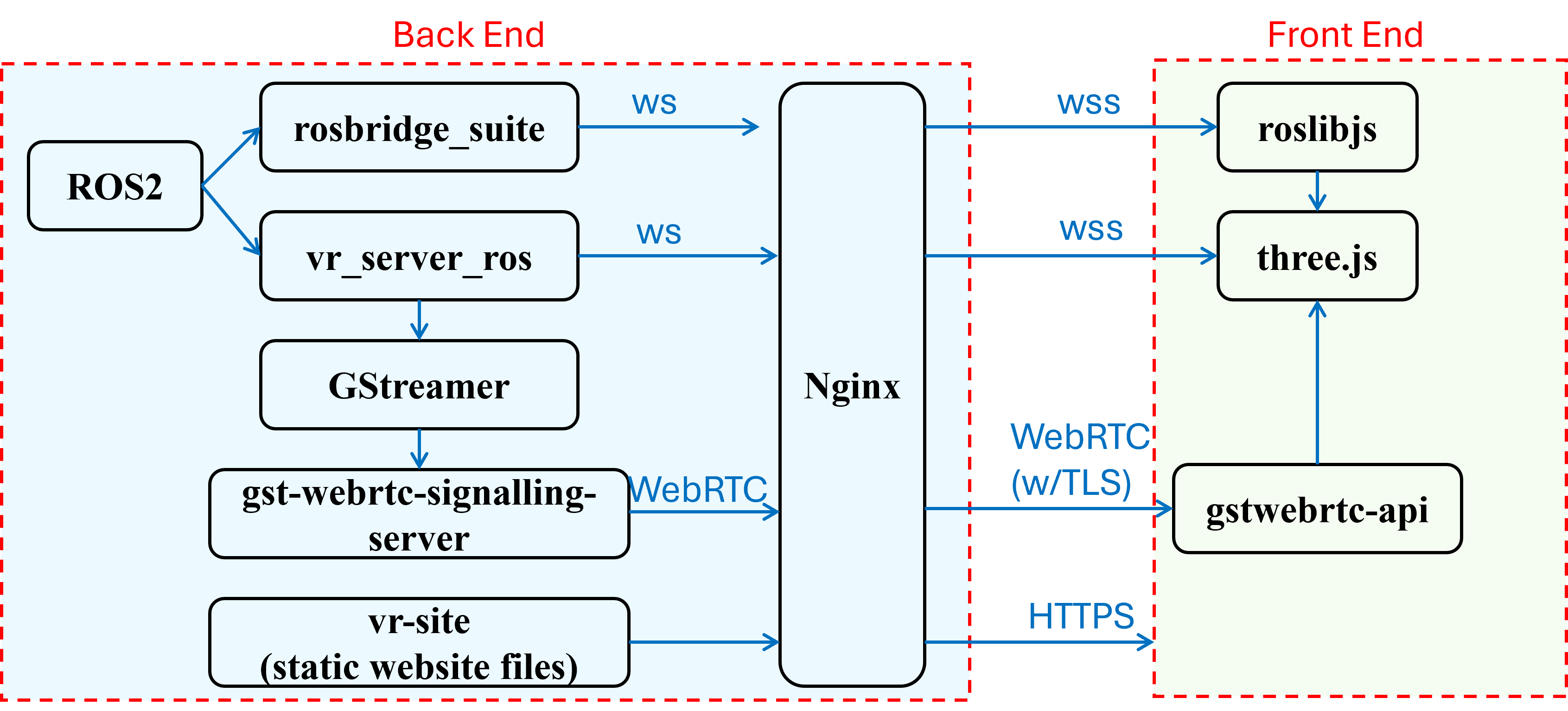}
    \caption{\label{softwaresetup}Functionality of software setup.}
\end{figure}

    Figure \ref{softwaresetup} illustrates our software architecture.  Three.js is used for rendering the VR scene on the front-end, which is done onboard the headset to minimize latency.  There are three main communication channels between the back-end and front-end: ROS2 communication, a websocket for mesh data, and a WebRTC video stream.  Most data, including control signals from the user and tracking data from the camera, are sent over ROS2 via the rosbridge suite.  We determined that sending the mesh data over ROS2 resulted in significantly reduced performance due to the need to serialize and deserialize the large mesh structures to and from ROS2 messages.  For this reason, the mesh data is sent in a custom binary format that minimizes serialization and deserialization costs.  Additionally, WebRTC is used to send a video stream because it is a low latency, low bandwidth solution.

    % talk about controller control
    To demonstrate the possibility of controlling the hexacopter exclusively over the WiFi connection using our system, a simple control scheme was implemented for the standard VR controllers included with the Quest 3 headset.  Pilots can interact with the right-hand joystick to control the velocity of the hexacopter in any direction in the XY plane.  For example, when the joystick is pushed upward, the hexacopter will move at a constant velocity in the forward direction.  
    
    An important statistic when evaluating a system's capability for real-time use is latency.  For example \cite{10.1007/978-3-319-58475-1_4} concludes that the latency threshold to be perceived as real time is significantly below 100 ms, and their experiments found that the median latency perception threshold is 54 ms.

    % \begin{oldblock}
    % To measure the latency of multi-copter tracking information, we used timestamped ROS2 messages, which are sent to the headset and echoed back to the VR server.  We can then divide the round-trip latency by two to get the one-way latency.  Rendering latency can be considered negligible, so this value can be considered an end-to-end latency measurement.
    % \end{oldblock}

% ======== Figures for Software Setup / Latency results (clean, deduped) ========

% Existing explanatory text above...

\begin{revblock}
\textbf{Latency measurement.}
At each frame, the VR server publishes a timestamped \texttt{PoseStamped} on \texttt{/vrserver/tracking}. 
The client echoes the most recent message back on \texttt{/vrserver/tracking\_echo}. 
We compute round-trip time (RTT) between transmit and echo receipt and report a one-way estimate as RTT$/2$. 
Rendering on the headset is small relative to transport and treated as negligible. 
One-way statistics (median, mean, P95, max, and fraction $>100$\,ms) are computed offline from the echo logs for pre/post comparisons.
\end{revblock}

% System diagram (single column)

% Commercial comparison (single column)
\begin{figure}[t]
    \centering
    \includegraphics[width=\linewidth]{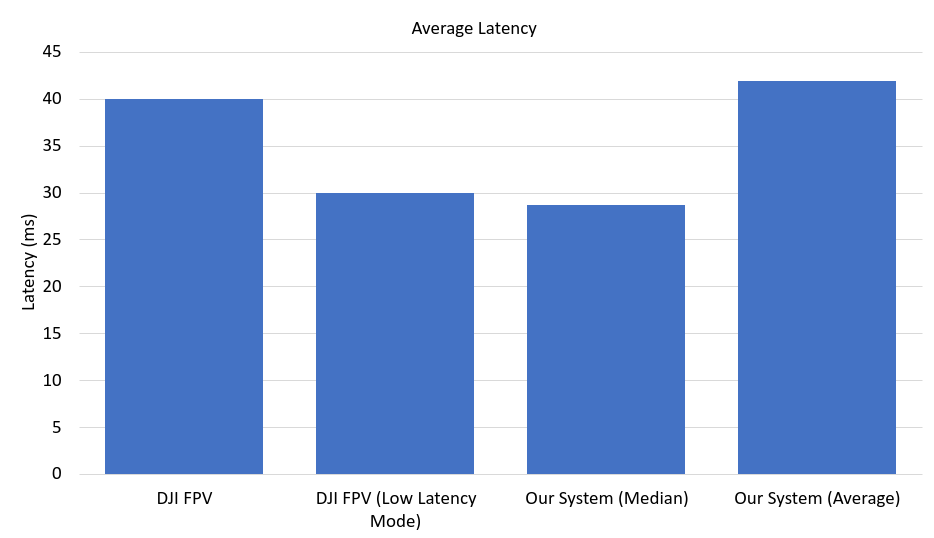}
    \caption{Comparison of latency with commercial offering.}
    \label{fig:latency-comparison}
\end{figure}

% Pre-optimization or representative trace (two-column)
\begin{figure*}[!t]
    \centering
    \includegraphics[width=0.8\paperwidth]{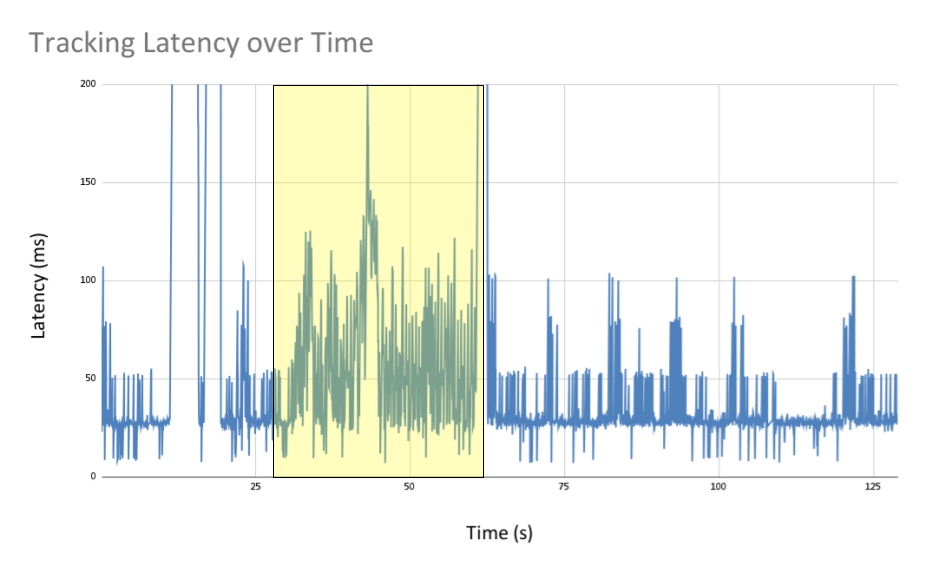}
    \caption{\revcap{One-way tracking latency over time estimated as RTT/2. The shaded interval marks a sustained mapping burst where variance rises. The two highest excursions occur at system startup and at a terminal emergency stop issued via the PX4 controller at the end of the flight. Operators can pause mapping or switch to FPV when tail latency grows (see Table~\ref{tab:latency}).}}
    \label{fig:latency}
\end{figure*}

% Post-optimization time series (two-column)
\begin{figure*}[!t]
    \centering
    \includegraphics[width=0.8\paperwidth]{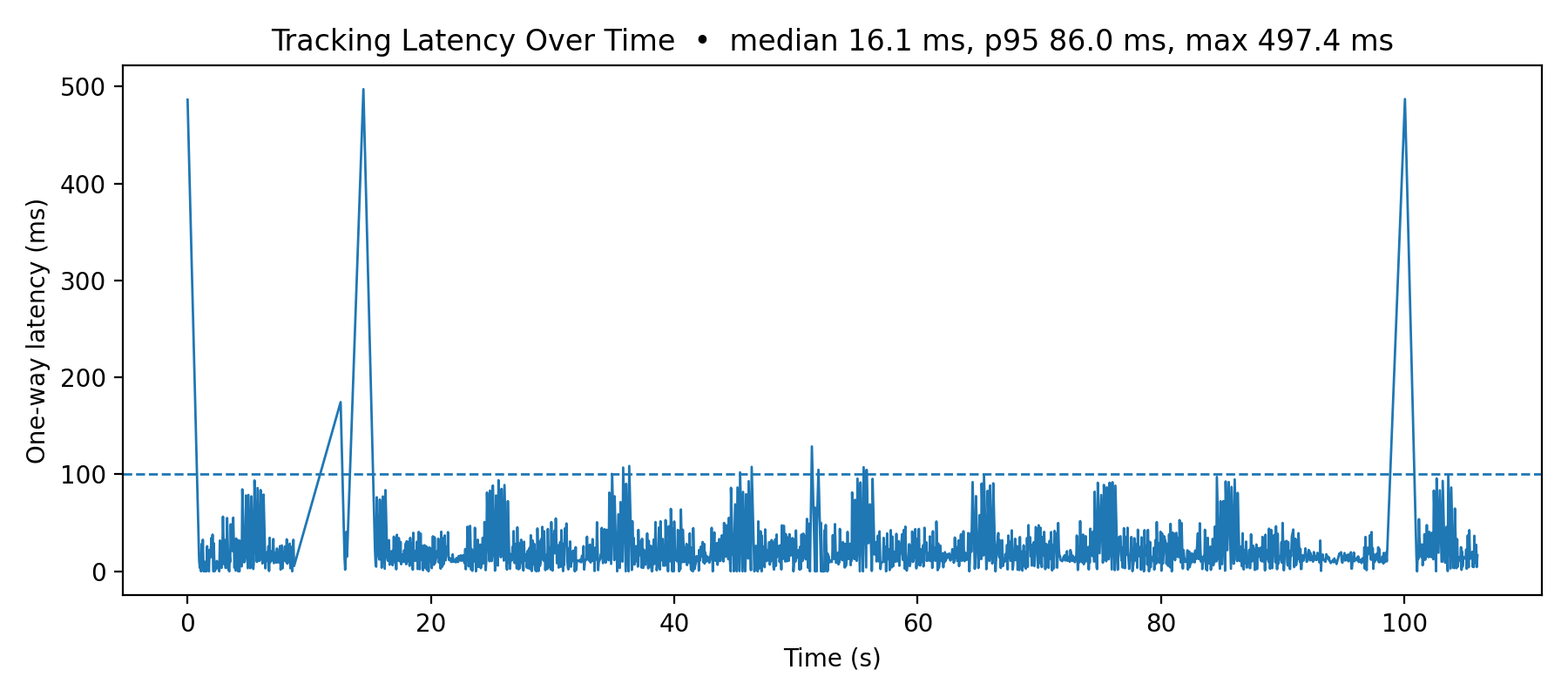}
   \caption{\revcap{Post-optimization one-way tracking latency over time (RTT/2). Dashed line shows 100\,ms. Full run: median \textbf{16.1}\,ms, mean \textbf{29.8}\,ms, P95 \textbf{86.0}\,ms, max \textbf{497.4}\,ms, and \textbf{3.16}\% of frames $>100$\,ms. Steady state, excluding the first 10\,s and frames $>250$\,ms: median \textbf{15.9}\,ms, mean \textbf{24.2}\,ms, P95 \textbf{74.6}\,ms, max \textbf{246.4}\,ms, and \textbf{1.47}\% $>100$\,ms. The two largest spikes are attributable to system startup and a terminal PX4 emergency stop; neither reflects steady-state operation.}}
    \label{fig:post_latency_timeseries}
\end{figure*}

% Compact summary (single column)
\begin{figure}[t]
    \centering
    \includegraphics[width=\linewidth]{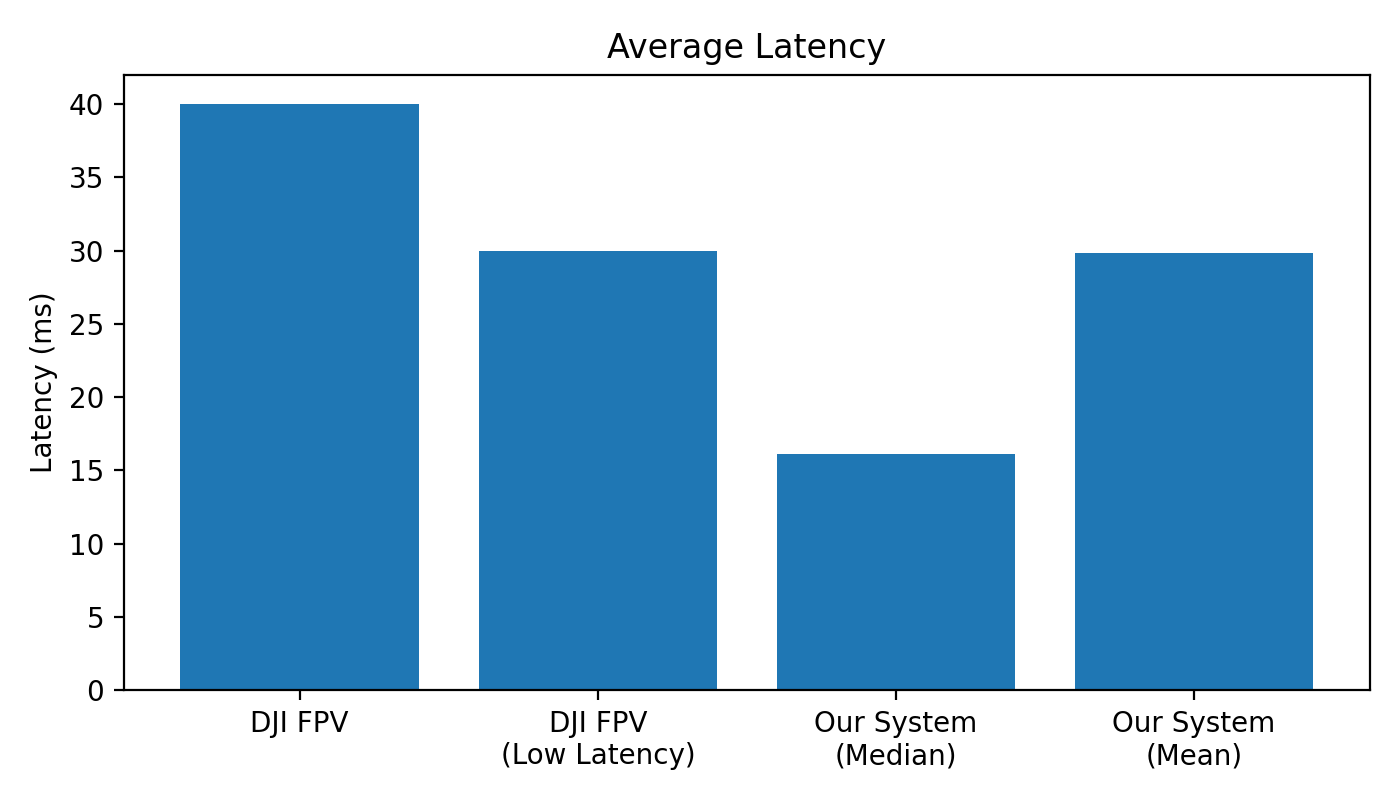}
    \caption{\revcap{Post-optimization latency summary. Full run: median \textbf{16.1}\,ms, mean \textbf{29.8}\,ms, P95 \textbf{86.0}\,ms, max \textbf{497.4}\,ms, and \textbf{3.16}\% $>100$\,ms. Steady state, excluding first 10\,s and frames $>250$\,ms: median \textbf{15.9}\,ms, mean \textbf{24.2}\,ms, P95 \textbf{74.6}\,ms, max \textbf{246.4}\,ms, and \textbf{1.47}\% $>100$\,ms.}}
    \label{fig:post_latency_summary}
\end{figure}

% Single-column: combined FPV vs TPV comparison (stacked panels)
\begin{figure}[t]
    \centering
    \subfloat[Task time]{%
        \includegraphics[width=\linewidth]{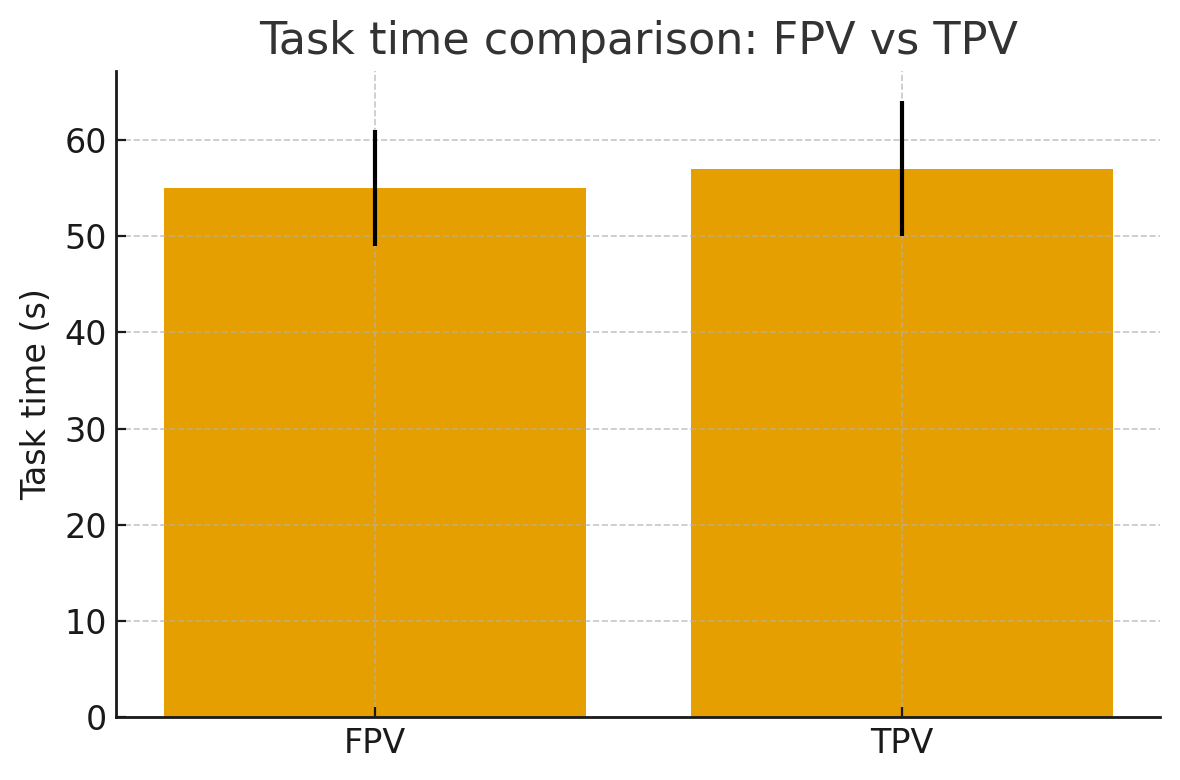}%
        \label{fig:tpvfpv_time_sub}}
    
    \subfloat[Minimum obstacle distance]{%
        \includegraphics[width=\linewidth]{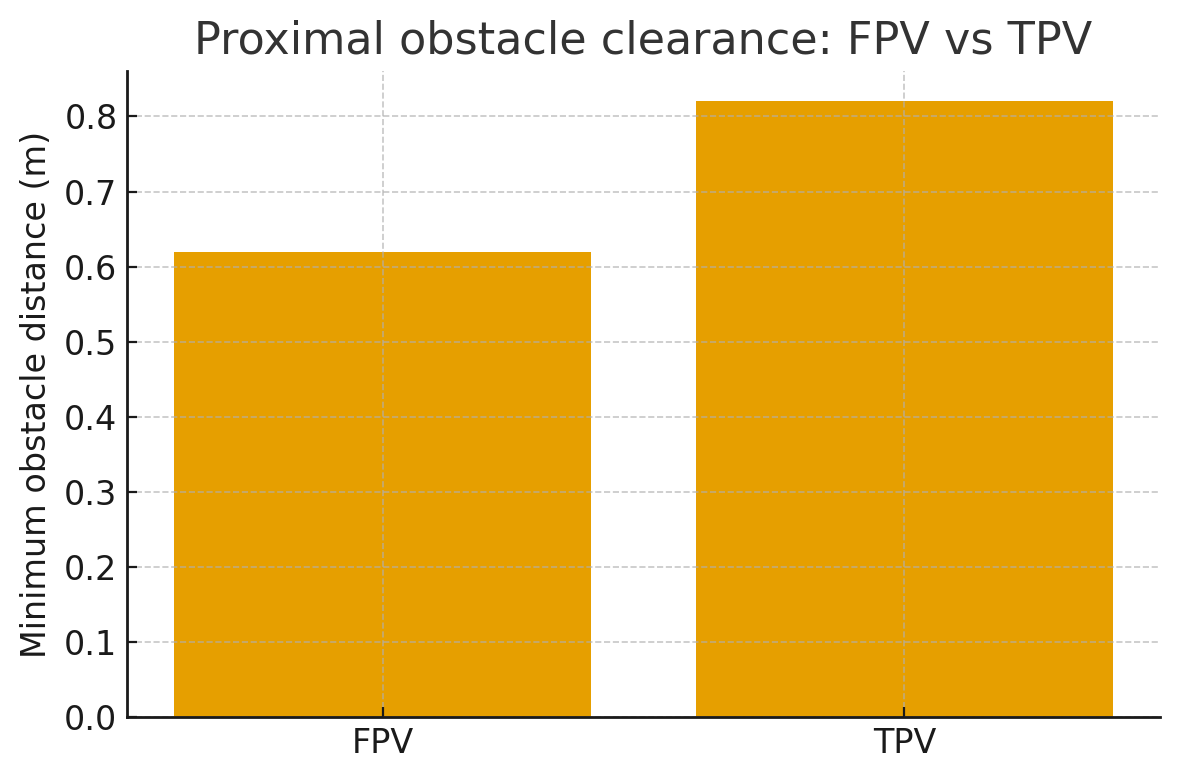}%
        \label{fig:tpvfpv_mindist_sub}}
    \caption{\revcap{FPV vs TPV performance in matched trials. Top, task time: FPV $55\pm6$\,s, TPV $57\pm7$\,s. Bottom, minimum distance: FPV 0.62\,m, TPV 0.82\,m.}}%
    \label{fig:tpvfpv_combined}
\end{figure}

% Single-column: environment screenshot (keep after comparisons)
\begin{figure}[t]
    \centering
    \includegraphics[width=\linewidth]{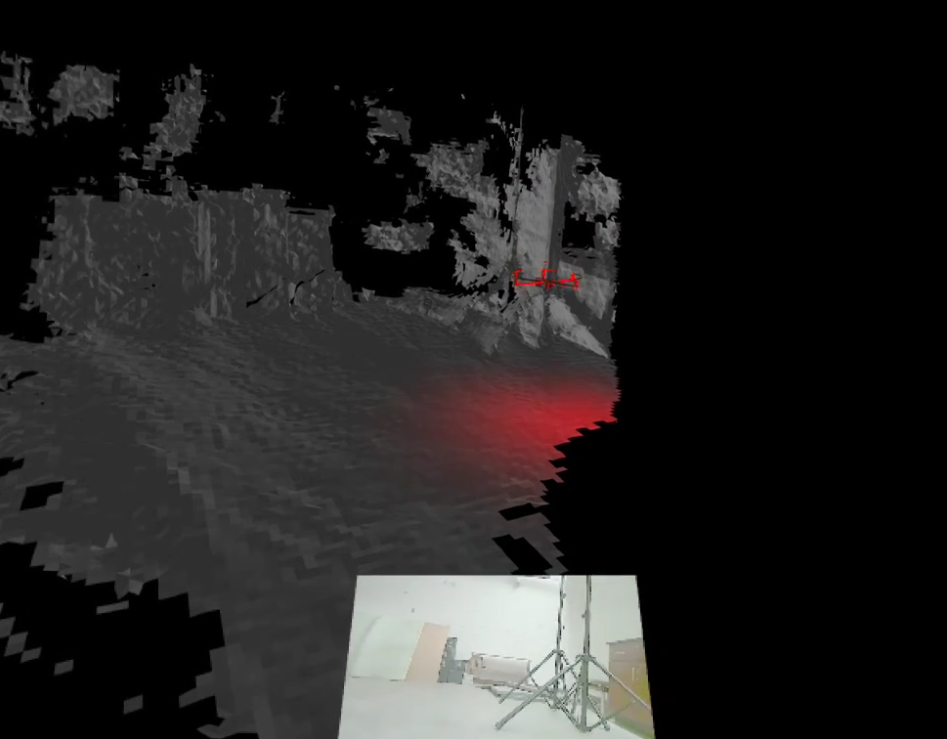}
    \caption{Screenshot of partially reconstructed environment, hexacopter, and video stream.}
    \label{fig:screenshot}
\end{figure}

% ======== End of figure block; text continues below ========

% The software paragraph and enumerated list follow here as you had them...
The software is all deployed on a Nvidia Jetson Orin NX. It is configured to launch the software at boot using a systemd service that runs a ROS2 launch file, and offers the following advantages:
\begin{enumerate}
    \item Portability and flexibility, WebXR works on any compatible headset.  
    \item Bandwidth utilization, compressed video and a binary mesh format reduce load. 
    \item Security, transport is encrypted with TLS.
\end{enumerate}

% Next section header continues as in your source:
\subsection{Robustness and Operational Fallbacks}
\label{sec:robustness}
\begin{revblock}
\textbf{Manual mapping throttle.}
When transient load increases, the operator can disable mapping using the built-in toggle (ROS2 service \texttt{/vrserver/mapping}, controller \texttt{select}). In this mode, only pose and video are streamed.

\textbf{FPV-only fallback.}
If persistent congestion or mapping load impairs readability, the client can run in FPV mode by launching the page with \texttt{?mode=fpv}. This shows a full-screen video feed while still rendering vehicle pose.

\begin{figure}[t]
    \centering
    \includegraphics[width=\linewidth]{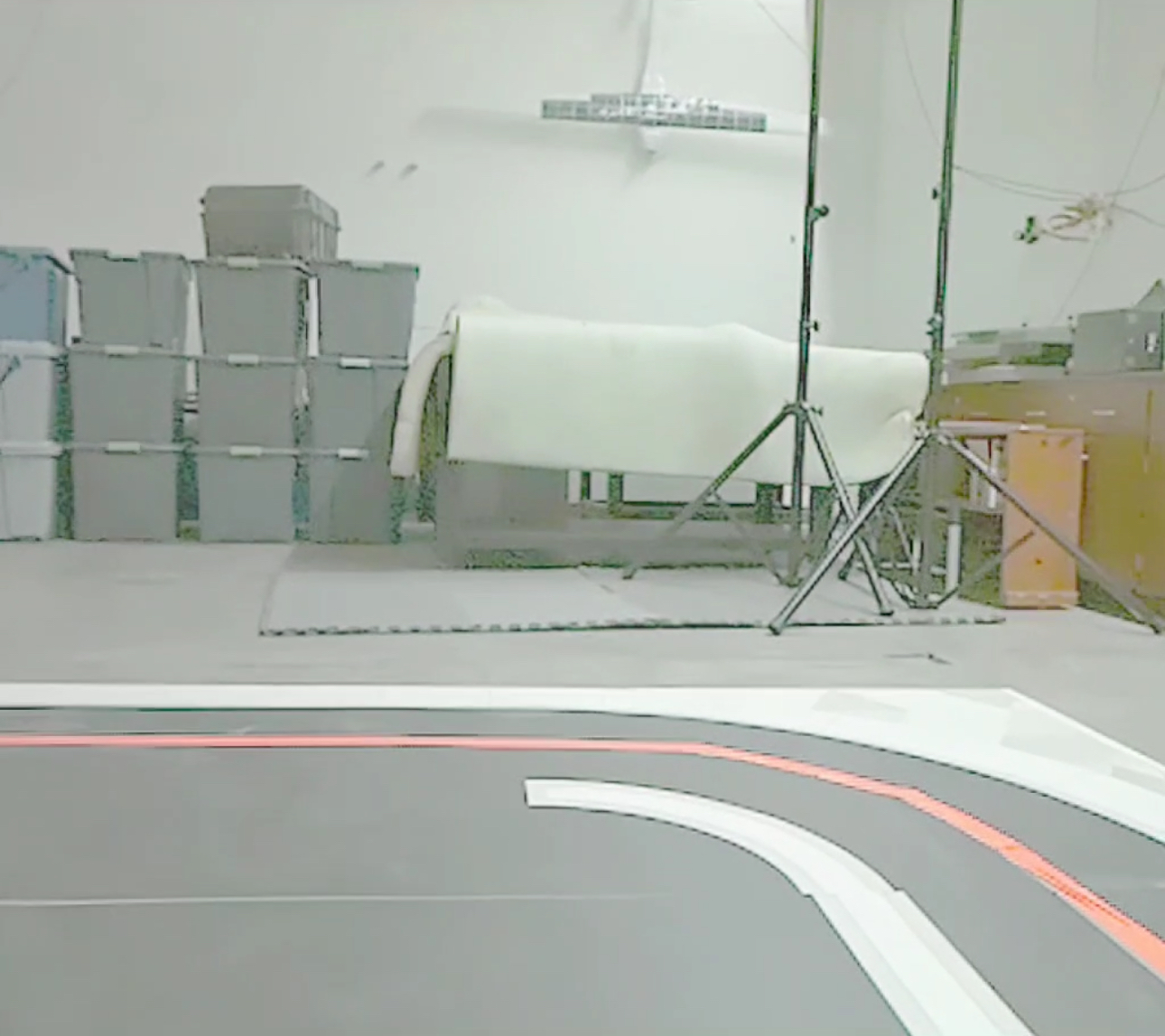}
    \caption{\revcap{FPV camera stream rendered in the VR client. FPV-only fallback prioritizes the video feed while maintaining pose cues.}}
    \label{fig:fpv_in_vr}
\end{figure}

\textbf{State indicators.}
The backend publishes tracking and mapping state on \texttt{/vrserver/status}; the front end surfaces status cues.
\textbf{Operator SOP.}
If one-way latency exceeds about 100\,ms or state is non-\texttt{OK} for more than about 2\,s, the sequence is: disable mapping, if needed switch to FPV, then re-enable mapping once stable.
\end{revblock}

\subsection{Limitations}

\begin{revblock}
% \textbf{Indoor-only evaluation and participant access.}
Our current evaluation is constrained to an indoor flight volume because the pose backbone relies on a Vicon motion-capture system. This infrastructure provides high-rate, high-accuracy ground truth but precludes outdoor trials, where GNSS availability, wind, lighting variability, and larger operational baselines may affect mapping quality and link behavior. The laboratory is a controlled facility with restricted access, which limits the participant pool. Two trained pilots flew the protocol. The sample size is small, so results should be interpreted as evidence of technical feasibility in controlled conditions rather than population-level usability. Future work will migrate to a VIO+GNSS/RTK outdoor stack and conduct an IRB-approved multi-participant study in open test ranges.
\end{revblock}

The system relies on the ZED camera and accompanying SDK for all simultaneous localization and mapping (SLAM) tasks, which limits the quality of the rendered environment to the quality that the ZED SDK can produce.  Real-time SLAM is a field that is rapidly advancing and future progress in that field will directly translate to an improved experience using systems such as this one as environment maps get more detailed.  Performance improvements in SLAM implementations will also lead to a significantly improved user experience because SLAM is the most compute-intensive task the system performs, and thus the main performance limitation.

\section{Experimental Results}

% \begin{oldblock}
% Our test pilot, while only interacting with the system, was able to take off, fly to a wall, stop and land safely, without having a direct visual line of sight to the hexacopter.  They were able to recognize the position of the wall by seeing the virtual representation of the wall that was scanned during the flight.  

% Furthermore, the average latency of the hexacopter tracking data during the duration of the test was 41.9 ms, with a median of 28.7 ms.  This means that in the virtual environment, the virtual drone's position and orientation are being accurately represented in real time, providing visual feedback to pilots equivalent to what they would receive through direct line of sight.  

% The system clearly demonstrated that the environment can be mapped and displayed to the pilot in real time, as shown in Fig. \ref{fig:screenshot} and the accompanying video. The measured latency between the system sending the data and it being displayed in the headset (see Fig. \ref{fig:latency-comparison}) had a median below commercial FPV drone offerings \cite{dji_fpv_specs}. Our system's average latency is significantly higher than the median due to significant latency spikes that can occur when mapping a large environment, as shown in the highlighted region in Fig. \ref{fig:latency}.  
% \end{oldblock}

\begin{revblock}
% \paragraph*{Participants, Task, and Conditions}
Two trained pilots completed the protocol. We evaluated \textbf{TPV} (third-person VR with live mapping) and an \textbf{FPV} baseline (video-only). Each pilot executed: takeoff $\rightarrow$ approach a wall and stop at commanded standoff $[\text{d}_{\mathrm{cmd}}]$ $\rightarrow$ return $\rightarrow$ land. For quantitative analysis we report repeated trials from one pilot ($N{=}8$) to control within-subject variability.

\noindent\textbf{Metrics.}
We logged task time, minimum obstacle distance, contacts, path length, and velocity smoothness (acceleration $\ell_2$). Latency was measured via timestamped ROS2 pings echoed from the headset, yielding one-way estimates (median, mean, P95, max, and fraction $>100$\,ms).

\noindent\textbf{TPV vs FPV Performance.} Table~\ref{tab:tpv-fpv} summarizes performance. Task time was comparable (FPV $55{\pm}6$\,s vs TPV $57{\pm}7$\,s). Minimum obstacle distance improved in TPV (FPV $0.62$\,m $\rightarrow$ TPV $0.82$\,m, $+0.20$\,m), and contacts were lower (FPV $1/4$ vs TPV $0/4$). Path length and smoothness differed by $<5\%$.

\begin{table}[t]
\centering
\caption{\revcap{TPV vs FPV (single pilot, repeated trials; reported metrics)}}
\label{tab:tpv-fpv}
\begin{tabular}{lrrrr}
\hline
Condition & Time (s) & Min Dist (m) & Contacts & Smoothness \\
\hline
FPV & $55{\pm}6$ & 0.62 & 1/4 & baseline \\
TPV & $57{\pm}7$ & 0.82 & 0/4 & within 5\% \\
\hline
\end{tabular}
\end{table}

\subsection{Latency after Optimizations}\label{sec:results_latency}
Using the timestamped echo method, the \emph{post-optimization one-way latency} for the full run is: median \textbf{16.1}\,ms, mean \textbf{29.8}\,ms, P95 \textbf{86.0}\,ms, max \textbf{497.4}\,ms, and \textbf{3.16}\% $>100$\,ms. 
Because the largest excursions occur at startup and at a terminal PX4 emergency stop, we also report a steady-state slice excluding the first 10\,s and frames $>250$\,ms: median \textbf{15.9}\,ms, mean \textbf{24.2}\,ms, P95 \textbf{74.6}\,ms, max \textbf{246.4}\,ms, and \textbf{1.47}\% $>100$\,ms.

\begin{table}[t]
\centering
\caption{\revcap{One-way tracking latency (ms), before vs. after optimizations}}
\label{tab:latency}
\begin{tabular}{lrrrrr}
\hline
Condition & Median & Mean & P95 & Max & \%$>$100 \\
\hline
Pre-opt (full) & 28.7 & 41.9 & 105.0 & 507.0 & 6.0 \\
Post-opt (full) & 16.1 & 29.8 & 86.0 & 497.4 & 3.16 \\
Post-opt (steady-state) & 15.9 & 24.2 & 74.6 & 246.4 & 1.47 \\
\hline
\end{tabular}
\end{table}

\noindent\textbf{Context vs Commercial Latency}
As in Fig.~\ref{fig:latency-comparison}, the post-optimization median remains below commonly cited real-time perception thresholds (Fig.~\ref{fig:post_latency_summary}) and comparable to commercial FPV offerings, while P95 and tail behavior improve when mapping bursts are avoided.
\end{revblock}

\section{Conclusion}
These results show that our system is a viable option for improving the spatial awareness of multi-copter pilots in an unknown environment.  The virtual environment, while rough, adequately informs pilots of nearby obstacles in a way that FPV cameras cannot.  The virtual drone mimics the movement of the real drone with a low enough latency for pilots to receive real-time visual feedback.  

Additionally, we have demonstrated that control can be achieved using VR controllers, sending control signals over WiFi.  This opens the door for fully remote operation of multi-copters with minimal loss in environmental awareness.

Other future work may include investigating alternative VR-based control options such as gestures, including additional cameras to map more of the environment at once, and attempting to add camera images as a texture on the rendered mesh to provide more detail about the environment.  
% - the possibility of adding textures on the renderened frames to give the world a colorful view.
\section{Acknowledgement}
 This work was supported by the
National Science Foundation under Award 2133690 and Award 1914581.

% \lipsum[1]%

% \section*{Supplemental Materials}
% \label{sec:supplemental_materials}

% Refer to the instructions for this section (\cref{sec:supplement_inst}).
% Below is an example you can follow that includes the actual supplemental material for this template:

% All supplemental materials are available on OSF at \url{https://doi.org/10.17605/OSF.IO/2NBSG}, released under a CC BY 4.0 license.
% In particular, they include (1) Excel files containing the data for and analyses for creating \cref{tab:vis_papers} and \cref{fig:vis_papers}, (2) figure images in multiple formats, and (3) a full version of this paper with all appendices.
% Our other code is intellectual property of a corporation---Starbucks Research---and there is no feasible way to share it publicly.

% \section*{Figure Credits}
% \label{sec:figure_credits}

% Refer to the instructions for this section (\cref{sec:figure_credits_inst}).
% Here are the actual figure credits for this template:

% \Cref{fig:teaser} image credit: Scott Miller / Special to the Vancouver Sun, January 22, 2009, page A6.

% \Cref{fig:vis_papers} is a partial recreation of Fig.\ 1 from \cite{Isenberg:2017:VMC}, which is in the public domain.

% \acknowledgments{This work was supported by the National Science Foundation under Award 2133690 and Award 1914581.}

% \bibliographystyle{abbrv}
% \bibliographystyle{abbrv-doi}
% \bibliographystyle{abbrv-doi-narrow}
% \bibliographystyle{abbrv-doi-hyperref}
% \bibliographystyle{abbrv-doi-hyperref-narrow}
% \clearpage
\bibliographystyle{IEEEtran}
\bibliography{template}

\begin{thebibliography}{10}
\providecommand{\url}[1]{#1}
\csname url@samestyle\endcsname
\providecommand{\newblock}{\relax}
\providecommand{\bibinfo}[2]{#2}
\providecommand{\BIBentrySTDinterwordspacing}{\spaceskip=0pt\relax}
\providecommand{\BIBentryALTinterwordstretchfactor}{4}
\providecommand{\BIBentryALTinterwordspacing}{\spaceskip=\fontdimen2\font plus
\BIBentryALTinterwordstretchfactor\fontdimen3\font minus \fontdimen4\font\relax}
\providecommand{\BIBforeignlanguage}[2]{{%
\expandafter\ifx\csname l@#1\endcsname\relax
\typeout{** WARNING: IEEEtran.bst: No hyphenation pattern has been}%
\typeout{** loaded for the language `#1'. Using the pattern for}%
\typeout{** the default language instead.}%
\else
\language=\csname l@#1\endcsname
\fi
#2}}
\providecommand{\BIBdecl}{\relax}
\BIBdecl

\bibitem{han2022uas}
Y.~Han, J.~Wei, and A.~Yilmaz, ``Uas navigation in the real world using visual observation,'' in \emph{2022 IEEE Sensors}.\hskip 1em plus 0.5em minus 0.4em\relax IEEE, 2022, pp. 1--4.

\bibitem{hobbs2016human}
A.~Hobbs and B.~Lyall, ``Human factors guidelines for unmanned aircraft systems,'' \emph{Ergonomics in Design}, vol.~24, no.~3, pp. 23--28, 2016.

\bibitem{sato2025development}
R.~Sato, E.~M. Badard, C.~S. Romulo, T.~Wada, and A.~Ming, ``Development of an aerial manipulation system using onboard cameras and a multi-fingered robotic hand with proximity sensors,'' \emph{Sensors}, vol.~25, no.~2, p. 470, 2025.

\bibitem{holuvsa2022utilization}
V.~Holu{\v{s}}a, F.~Bene{\v{s}}, and M.~Van{\v{e}}k, ``Utilization of augmented and virtual reality in geoscience,'' \emph{GeoScience Engineering}, vol.~68, no.~1, pp. 22--32, 2022.

\bibitem{kamran2024applications}
A.~Kamran-Pishhesari, A.~Moniri-Morad, and J.~Sattarvand, ``Applications of 3d reconstruction in virtual reality-based teleoperation: A review in the mining industry,'' \emph{Technologies}, vol.~12, no.~3, p.~40, 2024.

\bibitem{de2020virtual}
M.~E. de~Oliveira and C.~G. Corr{\^e}a, ``Virtual reality and augmented reality applications in agriculture: a literature review,'' in \emph{2020 22nd symposium on virtual and augmented reality (svr)}.\hskip 1em plus 0.5em minus 0.4em\relax IEEE, 2020, pp. 1--9.

\bibitem{huuskonen2018soil}
J.~Huuskonen and T.~Oksanen, ``Soil sampling with drones and augmented reality in precision agriculture,'' \emph{Computers and electronics in agriculture}, vol. 154, pp. 25--35, 2018.

\bibitem{cheng2022construction}
J.-Y. Cheng, E.~Lunardini Silva~Mendes, M.~Gheisari, and I.~Jeelani, ``Construction worker-drone safety training in a 360 virtual reality environment: A pilot study,'' \emph{EPiC Series in Built Environment}, vol.~3, 2022.

\bibitem{albeaino2022dronesim}
G.~Albeaino, R.~Eiris, M.~Gheisari, and R.~R. Issa, ``Dronesim: A vr-based flight training simulator for drone-mediated building inspections,'' \emph{Construction Innovation}, vol.~22, no.~4, pp. 831--848, 2022.

\bibitem{9974281}
H.~Stedman, B.~B. Kocer, M.~Kovac, and V.~M. Pawar, ``Vrtab-map: A configurable immersive teleoperation framework with online 3d reconstruction,'' in \emph{2022 IEEE International Symposium on Mixed and Augmented Reality Adjunct (ISMAR-Adjunct)}, 2022, pp. 104--110.

\bibitem{10.1145/3574131.3574432}
\BIBentryALTinterwordspacing
Y.~Luo, J.~Wang, Y.~Pan, S.~Luo, P.~Irani, and H.-N. Liang, ``Teleoperation of a fast omnidirectional unmanned ground vehicle in the cyber-physical world via a vr interface,'' in \emph{Proceedings of the 18th ACM SIGGRAPH International Conference on Virtual-Reality Continuum and Its Applications in Industry}, ser. VRCAI '22.\hskip 1em plus 0.5em minus 0.4em\relax New York, NY, USA: Association for Computing Machinery, 2023. [Online]. Available: \url{https://doi.org/10.1145/3574131.3574432}
\BIBentrySTDinterwordspacing

\bibitem{9981656}
B.~B. Kocer, H.~Stedman, P.~Kulik, I.~Caves, N.~Van~Zalk, V.~M. Pawar, and M.~Kovac, ``Immersive view and interface design for teleoperated aerial manipulation,'' in \emph{2022 IEEE/RSJ International Conference on Intelligent Robots and Systems (IROS)}, 2022, pp. 4919--4926.

\bibitem{9341037}
C.~Liu and S.~Shen, ``An augmented reality interaction interface for autonomous drone,'' in \emph{2020 IEEE/RSJ International Conference on Intelligent Robots and Systems (IROS)}, 2020, pp. 11\,419--11\,424.

\bibitem{CHEN2020105579}
\BIBentryALTinterwordspacing
Y.~Chen, B.~Zhang, J.~Zhou, and K.~Wang, ``Real-time 3d unstructured environment reconstruction utilizing vr and kinect-based immersive teleoperation for agricultural field robots,'' \emph{Computers and Electronics in Agriculture}, vol. 175, p. 105579, 2020. [Online]. Available: \url{https://www.sciencedirect.com/science/article/pii/S0168169920311479}
\BIBentrySTDinterwordspacing

\bibitem{10156426}
M.~Allenspach, T.~Kötter, R.~Bähnemann, M.~Tognon, and R.~Siegwart, ``Design and evaluation of a mixed reality-based human-robot interface for teleoperation of omnidirectional aerial vehicles,'' in \emph{2023 International Conference on Unmanned Aircraft Systems (ICUAS)}, 2023, pp. 1168--1174.

\bibitem{doi:10.1142/S2301385020500168}
\BIBentryALTinterwordspacing
Q.~Li, J.~P.~n. Queralta, T.~N. Gia, Z.~Zou, and T.~Westerlund, ``Multi-sensor fusion for navigation and mapping in autonomous vehicles: Accurate localization in urban environments,'' \emph{Unmanned Systems}, vol.~08, no.~03, pp. 229--237, 2020. [Online]. Available: \url{https://doi.org/10.1142/S2301385020500168}
\BIBentrySTDinterwordspacing

\bibitem{9833300}
A.~Chalvatzaras, I.~Pratikakis, and A.~A. Amanatiadis, ``A survey on map-based localization techniques for autonomous vehicles,'' \emph{IEEE Transactions on Intelligent Vehicles}, vol.~8, no.~2, pp. 1574--1596, 2023.

\bibitem{9635985}
M.~Zhao, X.~Guo, L.~Song, B.~Qin, X.~Shi, G.~H. Lee, and G.~Sun, ``A general framework for lifelong localization and mapping in changing environment,'' in \emph{2021 IEEE/RSJ International Conference on Intelligent Robots and Systems (IROS)}, 2021, pp. 3305--3312.

\bibitem{9526756}
W.~Liu, W.~Sun, and Y.~Liu, ``Dloam: Real-time and robust lidar slam system based on cnn in dynamic urban environments,'' \emph{IEEE Open Journal of Intelligent Transportation Systems}, pp. 1--1, 2021.

\bibitem{9381521}
Q.~Zou, Q.~Sun, L.~Chen, B.~Nie, and Q.~Li, ``A comparative analysis of lidar slam-based indoor navigation for autonomous vehicles,'' \emph{IEEE Transactions on Intelligent Transportation Systems}, vol.~23, no.~7, pp. 6907--6921, 2022.

\bibitem{app13179877}
\BIBentryALTinterwordspacing
L.~Wijayathunga, A.~Rassau, and D.~Chai, ``Challenges and solutions for autonomous ground robot scene understanding and navigation in unstructured outdoor environments: A review,'' \emph{Applied Sciences}, vol.~13, no.~17, 2023. [Online]. Available: \url{https://www.mdpi.com/2076-3417/13/17/9877}
\BIBentrySTDinterwordspacing

\bibitem{10.1145/3544548.3580681}
\BIBentryALTinterwordspacing
M.~Inoue, K.~Takashima, K.~Fujita, and Y.~Kitamura, ``Birdviewar: Surroundings-aware remote drone piloting using an augmented third-person perspective,'' in \emph{Proceedings of the 2023 CHI Conference on Human Factors in Computing Systems}, ser. CHI '23.\hskip 1em plus 0.5em minus 0.4em\relax New York, NY, USA: Association for Computing Machinery, 2023. [Online]. Available: \url{https://doi.org/10.1145/3544548.3580681}
\BIBentrySTDinterwordspacing

\bibitem{temma}
R.~Temma, K.~Takashima, K.~Fujita, K.~Sueda, and Y.~Kitamura, ``Third-person piloting: Increasing situational awareness using a spatially coupled second drone,'' 10 2019, pp. 507--519.

\bibitem{s17081720}
\BIBentryALTinterwordspacing
J.~J. Roldán, E.~Peña-Tapia, A.~Martín-Barrio, M.~A. Olivares-Méndez, J.~Del~Cerro, and A.~Barrientos, ``Multi-robot interfaces and operator situational awareness: Study of the impact of immersion and prediction,'' \emph{Sensors}, vol.~17, no.~8, 2017. [Online]. Available: \url{https://www.mdpi.com/1424-8220/17/8/1720}
\BIBentrySTDinterwordspacing

\bibitem{Medeiros2018_ThirdPersonBadForNavVR}
D.~Medeiros, R.~K. dos Anjos, D.~Mendes, J.~M. Pereira, A.~Raposo, and J.~A. Jorge, ``Keep my head on my shoulders!: Why third-person is bad for navigation in {VR},'' in \emph{Proceedings of the 24th ACM Symposium on Virtual Reality Software and Technology (VRST '18)}.\hskip 1em plus 0.5em minus 0.4em\relax New York, NY, USA: Association for Computing Machinery, 2018.

\bibitem{10.1007/978-3-319-58475-1_4}
V.~Forch, T.~Franke, N.~Rauh, and J.~F. Krems, ``Are 100 ms fast enough? characterizing latency perception thresholds in mouse-based interaction,'' in \emph{Engineering Psychology and Cognitive Ergonomics: Cognition and Design}, D.~Harris, Ed.\hskip 1em plus 0.5em minus 0.4em\relax Cham: Springer International Publishing, 2017, pp. 45--56.

\end{thebibliography}

\begin{IEEEbiography}[{\includegraphics[width=1in,height=1.25in,clip,keepaspectratio]{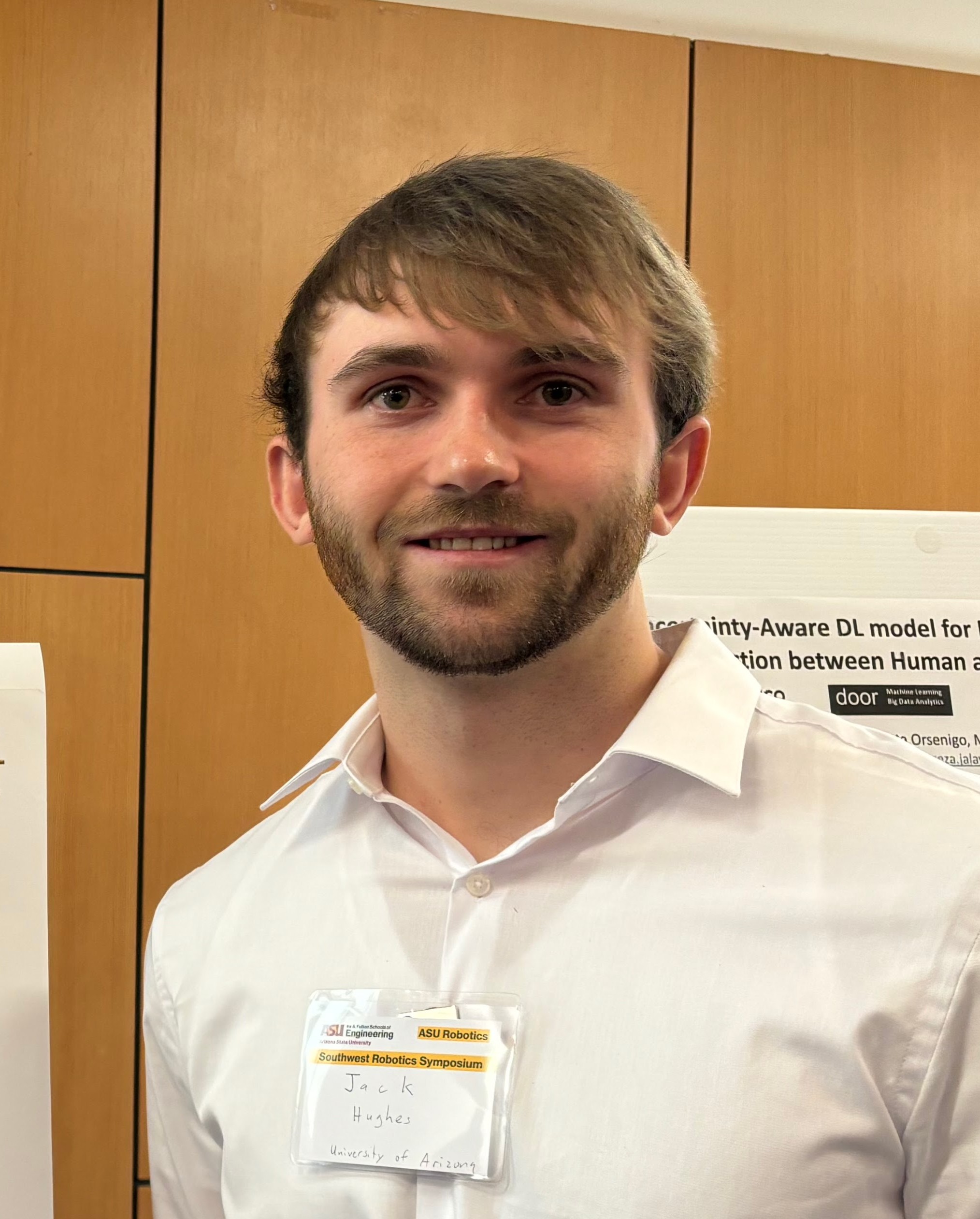}}]
{\textbf{Jack T. Hughes}} is  a Software Engineering major with a minor in Mathematics. His research interests include robotics, embedded software design, software optimization, and autonomous navigation and decision making. 
\end{IEEEbiography}

\begin{IEEEbiography}[{\includegraphics[width=1in,height=1.25in,clip,keepaspectratio]{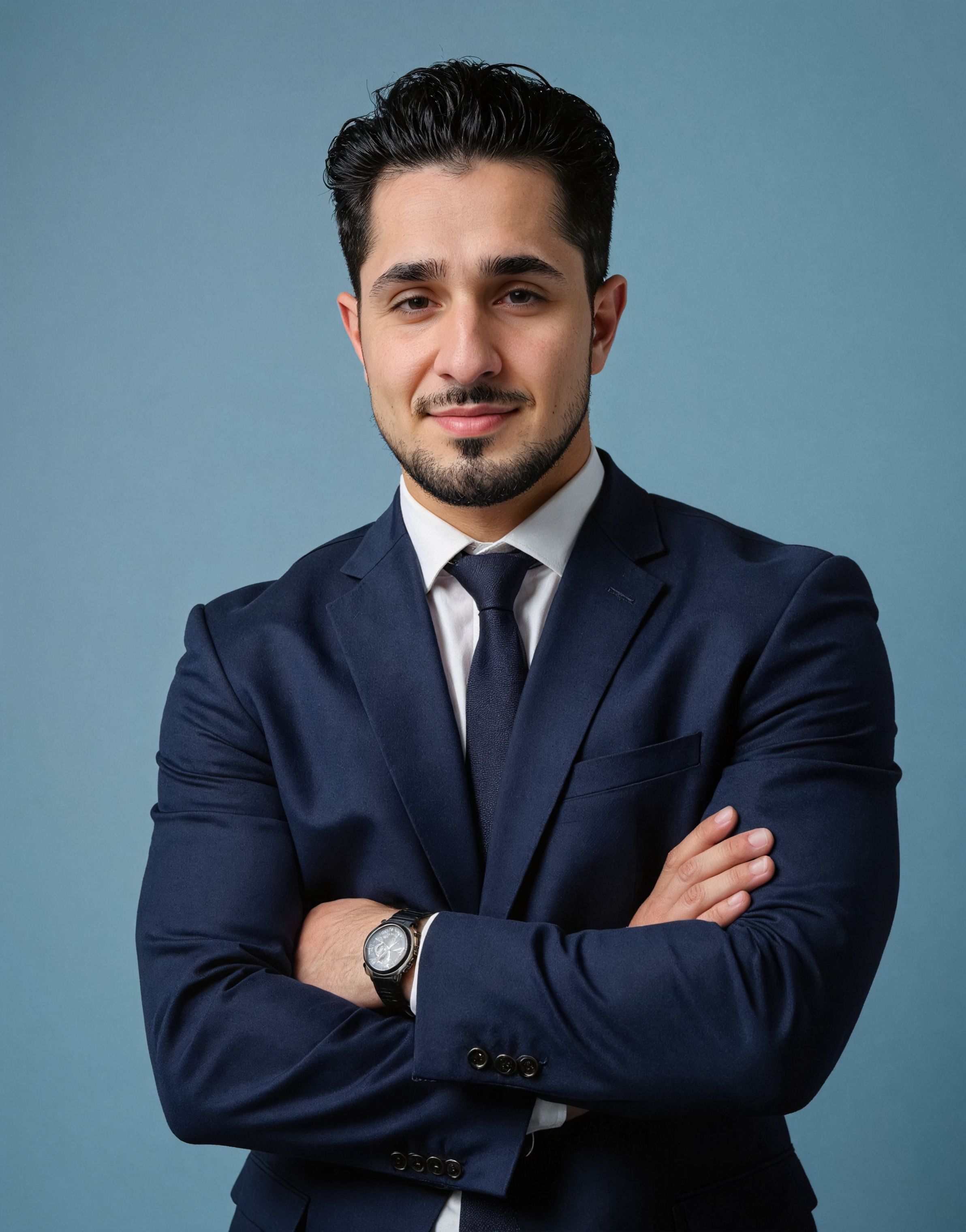}}]
{\textbf{Garegin Mazmanyan}}  holds a B.S. in Computer Science from the University of Arizona and is currently pursuing his M.S. in Computer Science at the same institution, specializing in artificial intelligence and robotics. His research focuses on Vision-Language-Action models for autonomous driving, emphasizing transformer-based control and deployment on Quanser’s QCar2 platform using ROS2. He has experience in real-time teleoperation, swarm coordination with Crazyswarm2, and VR-based immersive control of aerial and ground robots. His broader interests include multimodal learning, autonomy pipelines, and the development of safe and reliable robotic systems for real-world applications.
\end{IEEEbiography}

\begin{IEEEbiography}[{\includegraphics[width=1in,height=1.25in,clip,keepaspectratio]{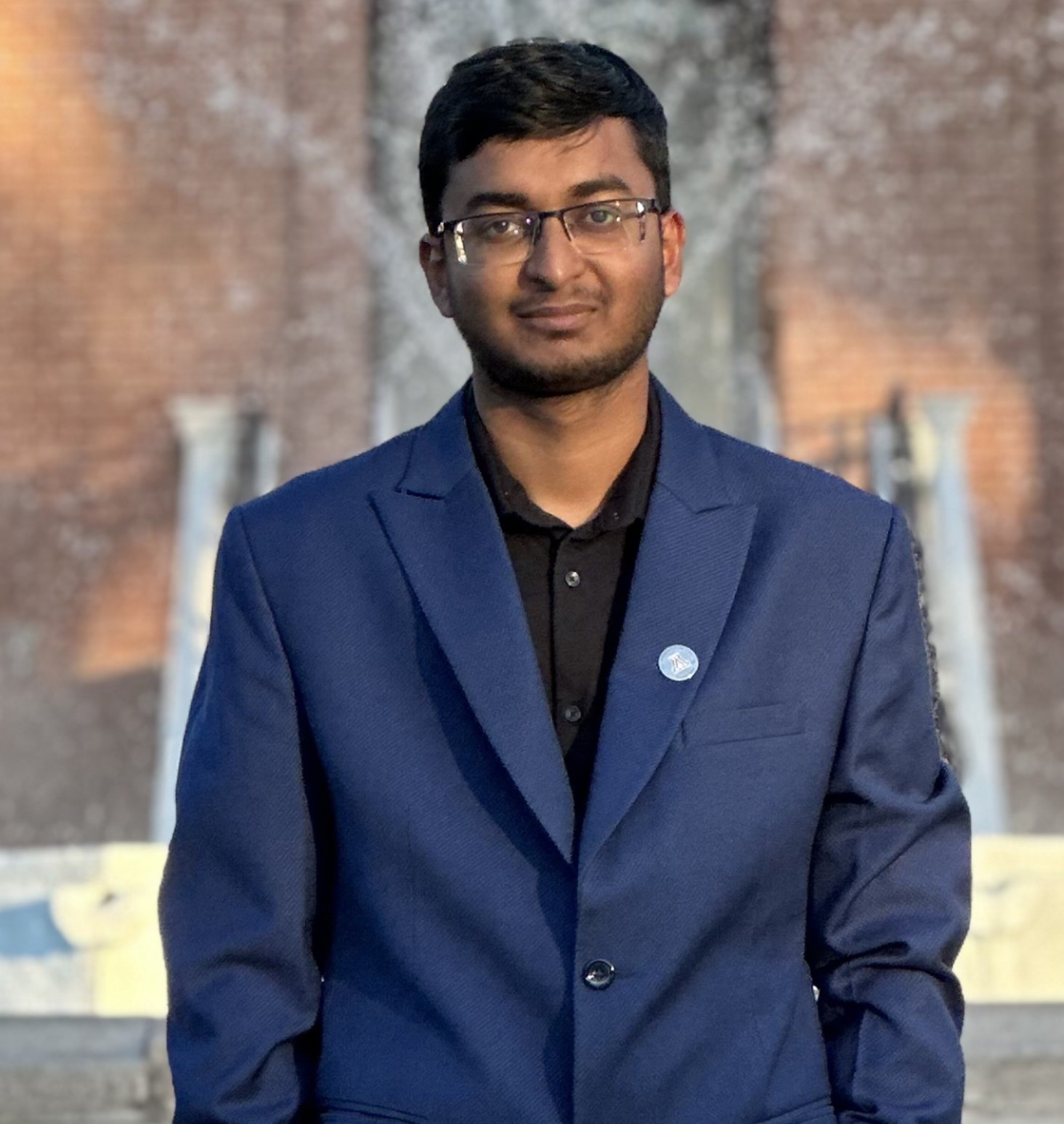}}]
{\textbf{Mohammad Ghufran}}  holds an M.Engg. in Robotics and Automation from the University of Arizona and a B.Tech. in Mechanical Engineering from India. His work focuses on robotics, mechatronics, sensor integration, and autonomous systems. His research experience spans motion planning, human-robot collaboration, and heterogeneous robot coordination. He is passionate about advancing intelligent automation by fusing mechanical design and control systems.
% Outside of academics, I play professional soccer for the University of Arizona and enjoy exploring new technologies.
% His current research interests include dynamics and control, multiagent systems, cyber-physical systems, and optimization and Markov decision processes.
\end{IEEEbiography}

\begin{IEEEbiography}[{\includegraphics[width=1in,height=1.25in,clip,keepaspectratio]{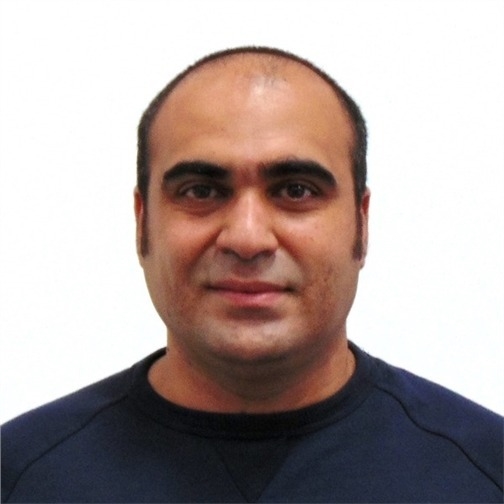}}]
{\textbf{Hossein Rastgoftar}} an Assistant Professor at the University of Arizona. Prior to this, he was an adjunct Assistant Professor at the University of Michigan from 2020 to 2021. He was also an Assistant Research Scientist (2017 to 2020) and a Postdoctoral Researcher (2015 to 2017) in the Aerospace Engineering Department at the University of Michigan Ann Arbor. He received the B.Sc. degree in mechanical engineering-thermo-fluids from Shiraz University, Shiraz, Iran, the M.S. degrees in mechanical systems and solid mechanics from Shiraz University and the University of Central Florida, Orlando, FL, USA, and the Ph.D. degree in mechanical engineering from Drexel University, Philadelphia, in 2015. 
% His current research interests include dynamics and control, multiagent systems, cyber-physical systems, and optimization and Markov decision processes.
\end{IEEEbiography}
\end{document}